\pdfoutput=1

\documentclass[11pt]{article}

\usepackage[final]{acl}

\usepackage{times}
\usepackage{latexsym}

\usepackage[T1]{fontenc}

\usepackage[utf8]{inputenc}

\usepackage{microtype}

\usepackage{inconsolata}

\usepackage{graphicx}

\title{The Mystery of In-Context Learning: A Comprehensive Survey on Interpretation and Analysis}
\author{Yuxiang Zhou$^{\heartsuit}$, Jiazheng Li$^{\heartsuit}$, Yanzheng Xiang$^{\heartsuit}$, Hanqi Yan$^{\heartsuit}$, Lin Gui$^{\heartsuit}$, Yulan He$^{\heartsuit\diamondsuit}$ \\
$^{\heartsuit}$Department of Informatics, King's College London~~~$^{\diamondsuit}$The Alan Turing Institute\\
        \texttt{\{yuxiang.zhou,jiazheng.li,yanzheng.xiang\}@kcl.ac.uk}\\
        \texttt{\{hanqi.yan,lin.1.gui,yulan.he\}@kcl.ac.uk}}

\usepackage{graphicx}
\usepackage{amsmath,bm}
\usepackage{amsfonts}
\usepackage{multirow}
\usepackage{booktabs}
\usepackage{microtype}
\usepackage{adjustbox}
\usepackage{tikz}
\usepackage{pgf}

\begin{document}
\maketitle

\begin{abstract}
Understanding in-context learning (ICL) capability that enables large language models (LLMs) to excel in proficiency through demonstration examples is of utmost importance.
This importance stems not only from the better utilization of this capability across various tasks, but also from the proactive identification and mitigation of potential risks, including concerns regarding truthfulness, bias, and toxicity, that may arise alongside the capability.
In this paper, we present a thorough survey on the interpretation and analysis of in-context learning.
First, we provide a concise introduction to the background and definition of in-context learning.
Then, we give an overview of advancements from two perspectives: 1) the theoretical perspective, emphasizing studies on mechanistic interpretability and delving into the mathematical foundations behind ICL; and 2) the empirical perspective, concerning studies that empirically analyze factors associated with ICL.
We conclude by discussing open questions and the challenges encountered and, by suggesting potential avenues for future research.
We believe that our work establishes the basis for further exploration into the interpretation of in-context learning. 
To aid this effort, we have created a repository\footnote{\url{https://github.com/zyxnlp/ICL-Interpretation-Analysis-Resources}} containing resources that will be continually updated.
\end{abstract}

\section{Introduction}
\begin{figure}
	\centering
	{\includegraphics[width=0.43\textwidth]{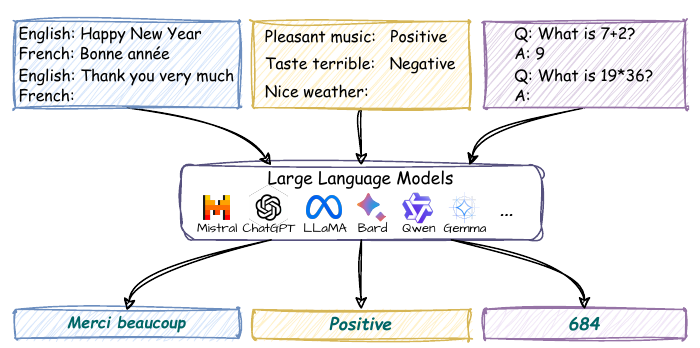}\label{vis(architure)}}
	\caption{Illustration of In-context Learning.}
 \label{fig:intro}
\end{figure}

The concept of in-context learning (ICL) was originally introduced by~\citet{Brown2020LanguageMA}, defined as `\textit{the model is conditioned on a natural language instruction and/or a few demonstrations of the task and is then expected to complete further instances of the task simply by predicting what comes next}'.
ICL is receiving increasing attention due to its remarkable adaptability and parameter-free nature.
As shown in Figure~\ref{fig:intro}, LLMs such as GPT-4~\cite{OpenAI2023GPT4TR}, Llama3~\cite{dubey2024llama}, and Qwen2~\cite{yang2024qwen2} have exhibited proficiency across various tasks, such as machine translation, sentiment analysis, and question answering, with a minimal set of task-oriented examples, all without re-training.
While ICL has been dominantly deployed in the Natural Language Processing (NLP) community, our understanding of ICL remains limited. 
Recently, an increasing number of studies have attempted to interpret and analyze ICL.
\citet{Garg2022WhatCT}, \citet{Dai2023WhyCG}, and \citet{Akyrek2023WHL} explained ICL through the lens of linear regression formulation.
\citet{Xie2021AnEO}, \citet{Wang2023LargeLM}, and \citet{Hahn2023ATO} provided an interpretation of ICL rooted in latent variable models.
Meanwhile, a distinct line of research has aimed to understand the influential factors affecting ICL through empirical analyses.
\citet{Min2022RethinkingTR}, \citet{Wei2023LargerLM}, \citet{Wang2023LabelWA}, and \citet{Kim2022GroundTruthLM} demonstrated that the ICL performance is influenced by task-specific characteristics and multiple facets of ICL instances, including quantities, order, and flipped labels. 
Consequently, it is essential to systematically categorize and summarize these studies, not only for a deeper understanding and more effective utilization of ICL across various tasks, but also to assist in anticipating and mitigating potential risks. 
These risks encompass concerns related to truthfulness, bias, and toxicity, that may arise alongside ICL.

In this paper, we present a thorough and organized survey of the research on the interpretation and analysis of ICL.
First, we provide a brief introduction of the background and offer the definition of ICL.
Then, we present a comprehensive overview of advancements, from two distinct viewpoints: 1) the \emph{theoretical perspective}, encapsulating studies focused on mechanistic interpretability and mathematical investigations into the  foundations of ICL; 
and 2) the \emph{empirical perspective}, pertaining to studies that prioritize empirical analysis by probing factors associated with ICL. 
In conclusion, we highlight the existing challenges and suggest potential avenues for further research. 
\section{Background and Notation}
In this section, we define the ICL paradigm using the following notations. A \textit{task} $\mathcal T_D$ consists of two components: a demonstration space~$\mathcal D$, which encompasses all possible demonstrations, and a joint probability distribution~$P_{(X, Y)}$.
A \textit{task demonstration} $D=\{(x_i,y_i)\}_{i=1}^{n} \in \mathcal D$ contains $n$ example pairs sampled from the joint distribution. In NLP, these example pairs could be Question-Answering pairs for QA tasks, parallel text for machine translation tasks, or sentence-label pairs for text classification tasks. For example, for machine translation, an example pair $(x, y)$ could be \textit{$($English: Happy New Year, French: Bonne année$)$}.
In contrast to traditional supervised learning approaches which aim to generalize from a fixed training dataset to predict future instances, ICL leverages continual exposure to demonstration examples and is guided by \textit{task query} to adapt the model dynamically to different contexts. 
In this survey, we denote the \textit{query} as $\mathcal T_Q =\{\mathcal Q, P_{Q}\}$, consisting a query space $\mathcal Q$ and marginal distribution $P_Q$.
A \textit{task query} $Q=\{q_j\}_{j=1}^m \in \mathcal Q$ contains $m$ instances sampled from the marginal distribution.
For example, $q$ could be ``\textit{English: Thank you French:}'' in the machine translation task.
Additionally, we define $A=\{a_j\}_{j=1}^m$ which represents the gold label for $Q$.
Let an LLM be defined as a function $F_\theta$ pre-trained on large-scale text corpora.
ICL can be defined as follows: 
\vspace{-1mm}
\paragraph{Definition (In-Context Learning)}\textit{In the context of a task query $Q$, in-context learning refers to the capability of $F_\theta$ to predict the correct answer $A$, conditioned on a task demonstration $D$.}

Based on the above definition, the process of ICL can be formally expressed as follows:
\begin{equation}
\begin{aligned}
 D \sim &P_{(X,Y)} \quad\quad Q \sim P_Q \\
& \hat{A} \leftarrow F_\theta(D,Q)
\end{aligned}
\end{equation}
The performance of ICL can be measured by: 
\begin{equation}
\vspace{-1mm}
 \mathcal S =  \mathbb E_{D,Q,A} \large[ M(\hat{A}, A)\large]
 \vspace{-1mm}
\end{equation}
$\hat{A}$ denotes the model-predicted output, and $M$ is an evaluation metric chosen based on the \textit{task query} $Q$ and its gold label $A$.

Based on our definition, we organize the existing literature on the interpretation and analysis of ICL into theoretical and empirical perspectives, as summarized in Table~\ref{tab:summ}. 
Researchers in the theoretical category focus on interpreting the connections among $F_\theta$, $D$, $Q$ and $A$ to explain the fundamental mechanism behind the ICL process.  
In contrast, those in the empirical category primarily centre on analyzing the relationship between performance $\mathcal S$ and the characteristics of the demonstration $D$.

\begin{table*}[t!]
  \centering
  \resizebox{0.9\linewidth}{!}{
 \setlength{\tabcolsep}{7mm}{
    \begin{tabular}{llll}\toprule
    \textbf{Work}  & \textbf{Key Words} & \textbf{Models} & \textbf{Tasks} \\
   		\midrule
   		\multicolumn{4}{c}{\textbf{Theoretical Perspective}}\\
    	\midrule 
        \citep{elhage2021mathematical}   & Mechanistic Interpretability & Transformer$^\dag$ & - \\
        \citep{Olsson2022IncontextLA}   & Mechanistic Interpretability & Transformer & - \\
        		\citep{Edelman2024TheEO}   & Mechanistic Interpretability & Transformer$^\dag$ & Markov Chain modeling \\
        		\citep{Swaminathan2023SchemalearningAR}   & Mechanistic Interpretability & Transformer & Next token prediction \\
        		\citep{Todd2024FunctionVI}   & Mechanistic Interpretability & Llama 2, GPT$^\ddag$ & Antonym, etc. \\
        		\citep{Bai2023TransformersAS}   & Mechanistic Interpretability &  Transformer$^\dag$ & Regression \\
        \citep{Garg2022WhatCT}   & Regression Function Learning & Transformer$^\dag$ & Regression \\
        \citep{Li2023TransformersAA}  & Regression Function Learning & Transformer$^\dag$ & Regression \\
        \citep{Li2023TheCO}   & Regression Function Learning & Transformer$^\dag$ & Regression \\
        \citep{Akyrek2023WHL}   & Regression Function Learning & Transformer$^\dag$ & Regression \\
        		\citep{Guo2024HowDT}   & Regression Function Learning & GPT$^\ddag$ & Representation Regression \\
        \citep{Dai2023WhyCG}   & Gradient Descent, Meta-Optimization & GPT$^\ddag$ & Classification \\
        \citep{Oswald2022TransformersLI}   & Gradient Descent, Meta-Optimization & Transformer$^\dag$ & Regression \\
                \citep{Oswald2023UncoveringMA}   & Gradient Descent, Meta-Optimization & Transformer$^\dag$ & Regression \\
                \citep{Deutch2024IncontextLA}   & Gradient Descent, Meta-Optimization & Transformer$^\dag$ & Classification \\
                \citep{Shen2024DoPT}   & Gradient Descent, Meta-Optimization &  LLaMa, GPT$^\ddag$ & Classification \\
                \citep{Fu2024TransformersLH}   & Gradient Descent, Meta-Optimization & Transformer, LSTM & Regression \\
        \citep{Xie2021AnEO}   & Bayesian Inference & Transformer$^\dag$, LSTM & Next token prediction  \\
        \citep{Wang2023LargeLM}   & Bayesian Inference & GPT$^\ddag$ & Classification \\
        \citep{Wies2023TheLO}   & Bayesian Inference & - & - \\
        \citep{Jiang2023ALS}  & Bayesian Inference & GPT$^\dag$&Sythetic Generation \\
        \citep{Zhang2023WhatAH}   & Bayesian Inference & Transformer$^\dag$ & - \\
        		\citep{Panwar2024InContextLT}   & Bayesian Inference & Transformer & Regression \\
        		\citep{Jeon2024AnIA}   & Bayesian Inference & Transformer$^\dag$ & Regression \\
        		\citep{Bigelow2024InContextLD}   & Bayesian Inference & GPT$^\ddag$ & Sequence generation \\

    	\midrule
    	\multicolumn{4}{c}{\textbf{Empirical Perspective}}\\
    	\midrule 
        \citep{Shin2022OnTE}   &  \textsc{Data} Domain & GPT-3 & Classification, etc. \\
        \citep{Han2023UnderstandingIL}   &  \textsc{Data} Domain, \textsc{Data} Distribution & OPT$^\ddag$ & Classification \\
        \cite{Raventos2023PretrainingTD} & Task Diversity & GPT-2 & Regression \\
        \citep{Razeghi2022ImpactOP}   &  \textsc{Data} Term frequency & GPT$^\ddag$ & Reasoning \\
        \citep{Kandpal2023LargeLM}   &  \textsc{Data} Term frequency & GPT-3$^\ddag$ & Question Answering \\
       	\citep{Chan2022DataDP}   & \textsc{Data} Distribution & Transformer & Classification \\
       	       	\citep{Yadlowsky2023PretrainingDM}   & \textsc{Data} Distribution & GPT-2$^\ddag$ & Regression \\
       	       	\citep{Hendel2023InContextLC}   & Task Diversity & LLaMA & Translation, etc. \\
        \citep{Wei2022EmergentAO}  & Model Scale & GPT-3$^\ddag$ & Classification \\
        \citep{Schaeffer2023AreEA}  & Model Scale, Evaluation Metric & GPT-3$^\ddag$ & Classification \\
        \cite{Tay2023UnifyingLL} & Pre-training Objective & UL2$^\ddag$ & Classification, etc. \\
        \citep{Kirsch2024GeneralPurposeIL} & Model Architecture & Transformer$^\dag$ & Classification \\
        		\citep{Singh2023TheTN} & Model Architecture & LLaMA $^\dag$ & Synthetic generation \\
        		\citep{Yousefi2024DecodingIL} & Embeddings & Llama, Vicuna & Regression \\
        		\citep{Akyrek2024InContextLL} & Model Architecture & Transformer, LSTM, etc. & Language learning \\
        \citep{Lu2021FantasticallyOP}   & Demonstration Order & GPT$^\ddag$ & Classification \\
        \citep{Liu2023LostIT} & Demonstration Order   & GPT-3.5$^\ddag$ & Question Answering \\
        \citep{Zhao2021CalibrateBU}   & Demonstration & GPT$^\ddag$ & Classification, IE, IR \\
        \citep{Liu2021WhatMG}   & Demonstration Order & GPT-3 & Classification, QA, etc. \\
        \citep{Min2022RethinkingTR}   & Input-Label Mapping & GPT$^\ddag$ & Classification, etc. \\
        \citep{Kim2022GroundTruthLM}   & Input-Label Mapping & GPT$^\ddag$ & Classification \\
        \citep{Wei2023LargerLM}   & Input-Label Mapping & GPT-3$^\ddag$ & Classification, QA, etc. \\
        \citep{Pan2023WhatIL}   & Input-Label Mapping &  GPT-3$^\ddag$ & Classification, etc. \\
        \citep{Lin2024DualOM}   & Input-Label Mapping & Llama, LSTM, etc. & Classification, etc. \\
        \citep{Kossen2023InContextLI}   & Input-Label Mapping &  LLaMA$^\ddag$ & Classification \\
        \citep{Tang2023LargeLM}   & Input-Label Mapping & GPT$^\ddag$ & Classification \\
        \citep{Si2023MeasuringIB}   & Input-Label Mapping & GPT-3$^\ddag$ & Classification, QA, etc. \\
        \citep{Wang2023LabelWA}   & Input-Label Mapping & GPT2-XL & Classification \\
    \bottomrule
    \end{tabular}}}
\caption{Summary of research studies on the interpretation of ICL. QA stands for Question Answering. \textsc{Data} refers to pre-training data. The symbol $^\dag$ denotes specifically designed models. The $^\ddag$ denotes that either various models or models from different families were used.}
  \label{tab:summ}%
\end{table*}%

\section{Theoretical Interpretation of ICL}
\vspace{-1mm}
\subsection{Mechanistic Interpretability}
With the goal of reverse-engineering components of LLMs models into more understandable algorithms, \citet{elhage2021mathematical} developed a mathematical framework to decompose operations within  Transformers~\cite{Vaswani2017AttentionIA} for explaining ICL.
They discovered that one-layer attention-only Transformers can perform very primitive ICL by assessing patterns (e.g., bigrams) from parameters.
Furthermore, they found that two-layer Transformers manifest a more general ICL capability using \textit{induction head}.
The induction heads are composed of attention heads that implement an algorithm to complete token sequences by copying and generating sequences that have occurred before.
Building on this foundation, \citet{Olsson2022IncontextLA} later investigated the internal structures responsible for ICL by analyzing the induction head in a full Transformer architecture.
They implemented circuits consisting of induction head and \textit{previous token head}, which copies information from one token to its successor.
Their study revealed a phase change occurring early in the training of LLMs of various sizes and found that circuits play a crucial role in implementing most ICL in LLMs. 
One pivotal insight from \citet{Olsson2022IncontextLA} presented comprehensive arguments supporting their hypothesis that induction heads may serve as the primary mechanistic source of ICL in a significant portion of LLMs, particularly those based on transformer architectures.

\textcolor{black}{
Later, \citet{Edelman2024TheEO} extended \citet{Olsson2022IncontextLA} by introducing a Markov Chain sequence modeling task, where demonstrations are sampled from a Markov chain.
They showed that transformers learn statistical induction heads to approach the Bayes-optimal by computing the correct posterior probability of the next token, given all previous occurrences of the prior token.
On the contrary, \citet{Swaminathan2023SchemalearningAR} adopted an alternative approach to elucidate the principles underpinning the mechanisms of ICL by studying clone-structured causal graphs (CSCGs), a sequence-learning model.  
They demonstrated that LLMs and CSCGS share similar mechanisms underlying ICL, which consist of: (a) learning template circuits for pattern completion, (b) retrieving relevant templates, and (c) rebinding novel tokens within the templates. 
Following on, \citet{Todd2024FunctionVI} measured causal mediators across a distribution of different tasks to identify the information transported by the attention heads in ICL.
They discovered \textit{function vectors} (FVs), a small number of attention heads transport
information of the demonstrations within the Transformer's hidden states during ICL.
\citet{Bai2023TransformersAS} unveiled a general mechanism, \textit{in-context algorithm selection}, to interpret ICL from a statistical viewpoint. 
They first demonstrated that transformers can implement a broad class of standard machine learning algorithms, such as least squares, ridge regression, and Lasso. 
Then, they theoretically demonstrated that transformers can adaptively select from these algorithms to learn more complex ICL tasks. 
}
\vspace{-2mm}
\subsection{Regression Function Learning}
Several research studies posited that the emergence of ICL can be attributed to the intrinsic capability of models to approximate regression functions for a novel task query $Q$ based on the task demonstration $D$.
\citet{Garg2022WhatCT} first formally defined ICL as a problem of learning functions and explored whether Transformers can be trained from scratch to learn simple and well-defined function classes, such as linear regression functions.
\textcolor{black}{To achieve this, they derived $D=\{(x_i,f(x_i))\}_{i=1}^n$ using a function $f$ sampled from a linear function class and trained models to predict the function value $A=\{f(q_j)\}_{j=1}^{m}$ for the corresponding $Q=\{q_j\}_{j=1}^{m}$.}
Their empirical findings revealed that Transformers exhibited ICL abilities, as they manifested to ``learn'' previously unseen linear functions from examples, achieving an average error comparable to that of the optimal least squares estimator.
Furthermore, \citet{Garg2022WhatCT} demonstrated that ICL can be applied to more complex function classes, such as sparse linear functions and decision trees.
They posited that the capability to learn a function class through ICL is an inherent property of the model $F_\theta$.
Later, \citet{Li2023TransformersAA} extended \citet{Garg2022WhatCT} to interpret ICL from a statistical perspective. They derived generalization bounds for ICL, considering two types of input examples: sequences that are independently and identically distributed (i.i.d.) and trajectories originating from a dynamical system.
They established a multitask generalization rate for both types of examples, addressing temporal dependencies by associating generalization to algorithmic stability and framing ICL as an algorithm learning problem. 
They found that Transformers can implement near-optimal algorithms on classical regression problems and proved that self-attention has favourable stability properties by quantifying the influence of individual tokens on one another.


At the same time, \citet{Li2023TheCO} took a further step from the work of~\cite{Garg2022WhatCT} to gain a deeper understanding of the role of the softmax unit within the attention mechanism of LLMs. 
They sought to mathematically interpret ICL based on the softmax regression formulation represented as $\min _\mathbf{x}||\left\langle\exp (A \mathbf{x}), \mathbf{1}_n\right\rangle^{-1} \exp (A \mathbf{x})-\mathbf{b}||_2$.
They established the upper bounds for data transformations effected by a single self-attention layer and theoretically demonstrated that LLMs perform ICL in a way that is highly similar to gradient descent (GD).
\textcolor{black}{ 
\citet{Akyrek2023WHL} took a different approach by delving into the process through which ICL learns linear functions, rather than analysing the types of functions that ICL can learn.
Through an examination of the inductive biases and algorithmic attributes inherent in Transformer-based ICL, they discerned that ICL can be understood in algorithmic terms, and linear learners within the model may essentially rediscover standard estimation algorithms.
They showed that trained in-context learners (ICLs) closely align with the predictors derived from GD, ridge regression, and exact least-squares regression.
While the transition between these predictors varies with model depth and training set noise, they will converge to Bayesian estimators at large hidden sizes and depths.
Additionally, \citet{Akyrek2023WHL} provided a theoretical proof demonstrating that Transformers can implement learning algorithms for linear models using GD and closed-form ridge regression.}

\textcolor{black}{
To understand ICL in a more complex scenario, \citet{Guo2024HowDT} studying ICL in the setting of \textit{learning with representations}.
They extended \citet{Garg2022WhatCT} to consider a more general task demonstration $D =\{(x_i, \Phi^{\star}(x_i))\}_{i=1}^{n}$ where the label depends on the instance $x$ through representation function $\Phi^{\star}(x)$ rather than $f$.
They theoretically demonstrated the existence of Transformers that can implement an approximately optimal ICL algorithm with mild depth and size.
Furthermore, \citet{Guo2024HowDT} trained Transformers to analyzing ICL  with various mechanisms, such as copying~\cite{Olsson2022IncontextLA} and \textit{post-ICL representation selection}~\cite{Bai2023TransformersAS}.
Their empirical results showed that lower layers transform the dataset and upper layers perform linear ICL.
}
\vspace{-1mm}
\subsection{Gradient Descent \& Meta-Optimization}
\label{sec:gd}
In the realm of gradient descent (GD), \citet{Dai2023WhyCG} adopted a perspective of viewing LLMs as meta-optimizers and interpreting ICL as a form of implicit fine-tuning. 
They first conducted a qualitative analysis of Transformer attention, representing it in a relaxed linear attention form, and identified a dual relationship between it and GD. 
Through a comparative analysis between ICL and explicit fine-tuning, \citet{Dai2023WhyCG} interpreted ICL as a meta-optimization process. They further provided evidence that the Transformer attention head possesses a dual nature similar to GD~\cite{Irie2022TheDF}, where the optimizer produces meta-gradients based on the provided examples for ICL through forward computation.
Concurrently, \citet{Oswald2022TransformersLI} also proposed a connection between the training of Transformers on auto-regressive objectives and gradient-based meta-learning formulations. 
They examined how Transformers define a loss function based on the given examples and the mechanisms by which Transformers assimilate knowledge using the gradients of this loss function. 
Their findings suggest that ICL may manifest as an emergent property, approximating gradient-based ICL within the forward pass of the model.

\textcolor{black}{
Following on, \citet{Oswald2023UncoveringMA} extended \citet{Oswald2022TransformersLI} to uncover underlying gradient-based mesa-optimization algorithms driving  model predictions by reverse-engineering autoregressive Transformers trained on sequence modeling tasks.
They showed that these models exhibit ICL capability enabled by the \textit{mesa-layer}, a novel attention layer that efficiently solves a least-squares optimization problem.
}
\textcolor{black}{
On the contrary, \citet{Deutch2024IncontextLA} revisited the hypothesis that GD approximates ICL and highlighted core issues in the evaluation metrics and baselines of~\citet{Dai2023WhyCG}.
Their findings suggested a weak correlation between ICL and GD and revealed major discrepancies in the flow of information throughout the model between ICL and GD.
In a similar vein, \citet{Shen2024DoPT} highlight that prior studies verify their hypothesis by training models explicitly for ICL, which differs from practical setups in real model training. 
They showed that the hand-constructed weights used in these studies possess properties that do not match those in real world scenarios.
Furthermore, they observed that ICL and GD have different sensitivities to the order in which they observe demonstrations in natural settings.
\citet{Fu2024TransformersLH} also presented evidence showing that Transformers learn to perform ICL by implementing a higher-order optimization rather than GD.
They theoretically demonstrated that Transformer circuits can efficiently implement Newton's method~\cite{gautschi2011numerical} and empirically showed that Transformers achieve the same convergence rate as Newton's method while being exponentially faster than GD. 
}
\vspace{-1mm}
\subsection{Bayesian Inference}
\citet{Xie2021AnEO} were the first to interpret ICL through the lens of Bayesian inference, positing that LLMs have the capability to perform implicit Bayesian inference via ICL.
Specifically, they synthesized a small-scale dataset to examine how ICL emerges in Transformer models during pre-training on text with extended coherence. 
Their findings revealed that Transformers are capable of inferring latent concepts to generate coherent subsequent tokens during pre-training. Additionally, these models were shown to perform ICL by identifying a shared latent concept among examples during the inference process. 
Their theoretical analysis confirms that this phenomenon persists even when there is a distribution mismatch between the examples and the data used for pre-training,  particularly in settings where the pre-training distribution is derived from a mixture of Hidden Markov Models (HMMs)~\cite{Baum1966StatisticalIF}. 

Following on,~\citet{Wang2023LargeLM} and \citet{Wies2023TheLO} expanded the investigation of ICL by relaxing the assumptions made by~\citet{Xie2021AnEO}.
\citet{Wies2023TheLO} assumed that there is a lower bound on the probability of any mixture component, alongside distinguishable downstream tasks with sufficient label margins.
They proved that ICL is guaranteed to happen when the pre-training distribution is a mixture of downstream tasks.
\citet{Wang2023LargeLM} posited that ICL in LLMs essentially operates as a form of topic modeling that implicitly extracts task-relevant information from examples to aid in inference.
They characterized the data generation process using a causal graph and imposed no constraints on the distribution or quantity of samples. 
Their theoretical investigations revealed that ICL can approximate the Bayes optimal predictor when a finite number of samples are chosen based on the latent concept variable.
At the same time, \citet{Jiang2023ALS} also introduced a novel latent space theory extending the idea of~\citet{Xie2021AnEO} to explain ICL in LLMs.
Instead of focusing on specific data distributions generated by HMMs, they delved into general sparse data distributions and employed LLMs as a universal density approximator for the marginal distribution, allowing them to probe these sparse structures more broadly.
They also demonstrated that ICL in LLMs can be ascribed to Bayesian inference operating on the broader sparse joint distribution of languages.

To shed light on the significance of the attention mechanism for ICL from a Bayesian view,~\citet{Zhang2023WhatAH} defined ICL as the task of predicting a response that aligns with a given covariate based on examples derived from a latent variable model. 
They demonstrated that certain attention mechanisms converge towards the conventional softmax attention as the number of examples goes to infinity. 
These attentions, due to their encoding of Bayesian Model Averaging (BMA) algorithm~\cite{Wasserman2000BayesianMS} within their structure, empower the Transformer model to perform ICL.
\textcolor{black}{
\citet{Panwar2024InContextLT} extended the previous setup in~\cite{Garg2022WhatCT,Akyrek2023WHL} by testing the Bayesian hypothesis for ICL over both linear and nonlinear function families.
They found that Transformers mimic the Bayesian predictor to perform ICL, including generalizing new function classes not seen during pre-training. 
Furthermore, they demonstrated that the simplicity bias in ICL arises from the pre-training distribution and provided empirical evidence that Transformers solve mixtures of tasks, suggesting the Bayesian perspective could offer a unified understanding of ICL.
}

Concurrently, \citet{Jeon2024AnIA} took a different approach to revisit ICL as Bayesian inference without restrictive assumptions by introducing information-theoretic tool.
They decomposed error for the Bayes optimal predictor into meta-learning error and intra-task error, and derived an in-context error upper bound of $log(N)/\tau$ for the sparse mixture of transformers, where $N$ is the number of
mixture components and $\tau$ is the in-context length.
To analyze ICL as Bayesian model selection in a practical setting, \citet{Bigelow2024InContextLD} modeled latent concepts evoked in LLMs by different contexts.
They adopted random binary sequences as context and examined dynamics of ICL by manipulating properties of the data,  such as sequence length, based on the cognitive science of human randomness perception.
They defined \textit{subjective randomness} to investigate model behaviour and demonstrated sharp phase changes, where LLMs suddenly shift from one pattern of behaviour to another during text generation, supporting the theories of ICL as model selection.
\vspace{-2mm}
\section{Empirical Analysis of ICL}
\vspace{-1mm}
\subsection{Pre-training Data}
There has been controversy among researchers regarding the effect of pre-training data properties on the performance of ICL.
To analyze the correlation between the domain of a corpus and ICL performance, \citet{Shin2022OnTE} evaluated LLMs pre-trained with subcorpora from diverse sources (e.g., blog, community website, news articles) within the HyperCLOVA corpus \cite{Kim2021WhatCC}.
They found that the corpus sources significantly influenced ICL performance; however, a pre-training corpus aligned with the downstream task's domain does not always guarantee competitive ICL performance.
For instance, while LLMs trained on subcorpora from blog posts exhibited superior ICL performance, a model trained on news-related dataset  did not sustain this superiority in ICL scenarios.
\citet{Han2023UnderstandingIL} found that the effectiveness of pre-training data for ICL is not necessarily tied to its domain relevance to downstream tasks.
By using MAUVAE score~\cite{Pillutla2021MAUVEMT} to quantify information divergence between pre-training data and target task data, 
they observed that pre-training data containing low-frequency tokens and long-tail information tend to have greater impact on ICL.
Conversely, \citet{Razeghi2022ImpactOP} and \citet{Kandpal2023LargeLM} identified a positive correlation between ICL performance and the term frequency within pre-training data, suggesting the memorization capabilities can significantly influence ICL.

By the manipulation of pre-training tasks to be a uniform distribution, \citet{Raventos2023PretrainingTD} identified a \textit{diversity threshold} - quantified by the number of tasks seen during pre-training - that indicates the emergence of ICL.
They empirically demonstrated that LLMs cannot perform a new task through ICL if the diversity of the pre-training task falls below the threshold.
\citet{Chan2022DataDP} have identified three critical distributional properties of pre-training data that drive ICL:
1) the training data exhibits \textit{bursty distribution}~\cite{Sarkar2005ABM}, where tokens appear in clusters rather than being uniformly distributed over time;
2) the marginal distribution across tokens is highly skewed, exhibiting a high prevalence of infrequently occurring classes, following a \textit{ZipFian distribution}~\cite{Zipf1949HumanBA};
3) the token meanings or interpretations are dynamic rather than fixed, where a token can have multiple interpretations (e.g., polysemy) or multiple tokens may correspond to the same interpretation.

\textcolor{black}{
\citet{Yadlowsky2023PretrainingDM} explored the impact of pre-training data composition on the ability ICL.
Building on the setup of~\citet{Garg2022WhatCT}, they showed empirical evidence that the Transformers can perform model selection among pre-trained function classes during ICL with minimal additional cost.
However, there was no evidence that the models were able to generalize beyond their pre-training data through ICL.
To shed light on the mechanisms of ICL, \citet{Hendel2023InContextLC} investigated the relationship between demonstrations and the parameters of functions in certain hypothesis classes by examining the top tokens in the output distribution. 
They revealed that ICL functions by compressing training data into a task vector, which then guides Transformers to generate outputs.
}
\vspace{-1mm}
\subsection{Pre-training Model}
The attributes of pre-training models have been shown to be significant factors affecting ICL. 
\citet{Wei2022EmergentAO} focused on training computation (e.g., FLOPs~\cite{Hoffmann2022TrainingCL}) and model size(e.g, number of model parameters).
They analyzed the emergent manner in which ICL manifests with the scaling of LLMs and highlighted the positive correlation between the model's scale and ICL performance.
On the contrary, \citet{Schaeffer2023AreEA} empirically analyzed the effect of the choice of evaluation metric on the emergence of ICL ability.
By controlling for factors such as downstream task, model family, and model outputs, they found that this ability appears due to the choice of metric, rather than as a result of fundamental changes in models with scaling.
At the same time, \citet{Tay2023UnifyingLL} have suggested that the pre-training objective is a pivotal factor influencing ICL performance.
They observed that continued pre-training with varied objectives enables robust ICL performance.
\citet{Kirsch2024GeneralPurposeIL} have posited that aspects of model architecture, such as the dimension of the hidden size, play a more critical role than model size in the emergence of ICL.

\textcolor{black}{
\citet{Singh2023TheTN} suggested that the emergence of ICL should be viewed as a \textit{transient} rather than a \textit{persistent} phenomenon.
They demonstrated that ICL may not persist as the model continues to be trained.
By examining model sizes, pre-training data size, and domain, they found that ICL first emerges, then disappears, giving way to in-weights learning (IWL).
\citet{Yousefi2024DecodingIL} introduced a neuroscience-inspired framework to empirically analyze how LLM embeddings and attention representations change following ICL.
 They measured the ratio of attention information over parameterized probing classifiers based on representational similarity analysis (RSA) and found a meaningful correlation between improvements in behaviour after ICL and changes in both embeddings and attention weights across LLM layers.
 \citet{Akyrek2024InContextLL} proposed a novel dataset, REGBENC, to systematically study in context language learning (ICLL) in the setting of regular languages, the class of formal languages generated by finite automata~\cite{Hopcroft1971AnNL}. They investigated ICLL in relation to model classes and mechanisms by examining essential features such as structured outputs, probabilistic predictions, and compositional reasoning about input data.
 They found that Transformers are the most efficient and can develop higher-order variants of induction heads~\cite{Olsson2022IncontextLA}.
}
\subsection{Demonstration Order}
The order of the demonstrations has a significant impact on the ICL.
\citet{Lu2021FantasticallyOP} designed demonstrations containing four samples with a balanced label distribution and conducted experiments involving all 24 possible permutations of sample orders.
Their experimental results demonstrated that the ICL performance varies across different permutations and model sizes.
In addition, they found that effective prompts are not transferable across models, indicating that the optimal order is model-dependent, and what works well for one model does not guarantee good results for the other models.
Both \citet{Zhao2021CalibrateBU} and \citet{Liu2023LostIT} identified a similar phenomenon where LLMs tend to repeat answers found at the end of provided demonstrations in ICL.
Their results indicated that ICL performs optimally when the relevant information is positioned at the beginning or end of the demonstrations and the performance degraded when the LLMs are compelled to use information from the middle of the input.
\citet{Liu2021WhatMG} delved deeper and analyzed the underlying reasons for how the order of demonstration influences ICL.
They proposed retrieving examples semantically similar to a test example for creating its demonstration and found that the demonstration order appears to be dependent on the specific dataset in use.
\subsection{Input-Label Mapping}
Some studies have explored the impact of input-label mappings on the performance of ICL in LLMs.
\citet{Min2022RethinkingTR} empirically showed that substituting the ground-truth labels in demonstrations with random ones results in a marginal performance decrease across various tasks.
This indicates that ICL exhibits a low sensitivity to the accuracy of labels in the demonstration.
This finding contradicts the conclusions in ~\citet{Kim2022GroundTruthLM}, \citet{Wei2023LargerLM}, and \citet{Kossen2023InContextLI}, who argued that LLMs rely significantly on accurate input-label mappings to perform ICL.
For example, \citet{Kim2022GroundTruthLM} highlighted that averaging performance across multiple datasets fails to accurately reflect the insensitivity observed within specific datasets. 
They introduced two novel metrics \textit{label correctness sensitivity} and \textit{ground-truth label effect ratio}, to extensively quantify the impact of ground-truth labels on ICL performance.
Their empirical findings confirmed that ground-truth label significantly influences ICL, revealing a strong correlation between sensitivity to label correctness and the complexity of the downstream task.

To investigate the effect of semantic priors and input-label mappings on ICL, \citet{Wei2023LargerLM} discovered that LLMs can prioritize input-label mappings from demonstrations over pre-training semantic priors, leading LLMs to drop below random guessing when all the labels in the demonstrations are flipped.
Additionally, their research indicated that smaller models predominantly utilize the semantic meanings of labels rather than the input-label mappings presented in ICL demonstrations. 
\textcolor{black}{
\citet{Pan2023WhatIL} investigated how ICL leverages demonstrations by characterizing \textit{task recognition} and \textit{task learning} in LLMs.
They reported that LLMs exhibit a significantly better ability to learn input-label mappings through ICL when compared to smaller models. 
Moreover, they observed that the ability of ICL to discern tasks from demonstrations does not substantially improve with increased model size.
Following on, \citet{Lin2024DualOM} extended \citet{Pan2023WhatIL} by introducing multiple task groups and task-dependent input distributions to investigate the factor of pre-training data. 
They showed that ICL demonstrations with biased labels contain sufficient information to retrieve a correct pretrained task.
}
\citet{Tang2023LargeLM} revealed that LLMs may rely on shortcuts in ICL demonstrations for downstream tasks.
These shortcuts consist of spurious correlations between ICL examples and their associated labels.

\citet{Si2023MeasuringIB} identified that LLMs exhibited feature bias when provided with \textit{underspecified} ICL demonstrations in which two features are equally predictive of the labels.
Their experiment suggested that interventions such as employing instructions or incorporating semantically relevant label words could effectively mitigate bias in ICL.
To understand the underlying mechanism behind LLMs performing ICL from an information flow perspective, \citet{Wang2023LabelWA} discovered that semantic information is concentrated within the representations of label words in the shallow computation layers.
Furthermore, they showed that the consolidated information within label words acts as a reference for LLMs' final predictions, highlighting the importance of label words in ICL.
\vspace{-1mm}
\section{Open Questions}
\vspace{-1mm}
\textcolor{black}{
Despite ongoing endeavours to interpret and analyse ICL, we are still far from fully understanding it due to the open-ended nature of some questions.
For example, while \citet{elhage2021mathematical} and \citet{Olsson2022IncontextLA} contribute to our understanding of ICL by probing the internal architecture of LLMs, it is important to note that their findings represent initial steps towards the comprehensive reverse-engineering of LLMs. 
It becomes particularly intricate when dealing with LLMs characterized by complex structures comprising hundreds of layers and spanning billions to trillions of parameters.
This complexity introduces significant challenges.
Similarly, although studies have provided theoretical proofs and empirical evidence on the relation of ICL and regression function learning~\cite{Garg2022WhatCT,Akyrek2023WHL,Li2023TheCO,Li2023TransformersAA}, gradient descent or meta-optimization~\cite{Oswald2022TransformersLI,Dai2023WhyCG}, and Bayesian inference~\cite{Xie2021AnEO,Wang2023LargeLM,Wies2023TheLO,Jiang2023ALS, Zhang2023WhatAH}, their conclusions are limited to simplified model architectures and controlled synthetic experimental settings. 
This raises the open question of whether these findings hold in the context of standard model architectures without approximations and if they can be directly applied to real-world scenarios.
}

\textcolor{black}{
On the other hand, there are contradictory findings from researchers analysing the factors affecting ICL.
For instance, while \citet{Razeghi2022ImpactOP} and \citet{Kandpal2023LargeLM} identified a positive relation between ICL performance and term frequency in pre-training data, \citet{Han2023UnderstandingIL} found that pre-training data with long-tail and rarely occurring tokens contribute more significantly to ICL.
Additionally, while \citet{Min2022RethinkingTR} suggested that ICL exhibits low sensitivity to labels in the demonstrations, other studies revealed that accurate input-label mappings play an important role in performing ICL~\cite{Kim2022GroundTruthLM,Wei2023LargerLM,Kossen2023InContextLI}.
One of the core challenges in analysing ICL empirically lies in the necessity of controlling for numerous relevant variables, leading most existing conclusions to typically rely on correlations rather than causal relations. 
This raises open questions about the reliability of these findings and the extent to which confounding factors may influence the results.
}
\section{Future Directions}
\vspace{-1mm}
\paragraph{Correlation vs Causation} 
Most existing studies have interpreted ICL in LLMs primarily through correlational analyses, leading to biased conclusions that may not be broadly applicable.
One core challenge lies in various underlying factors that interact with each other and influence ICL~\cite{Wei2022EmergentAO, Lu2023AreEA}.
A potential approach involves designing qualitative and quantitative analysis for ICL by systematically accounting for a range of potential factors.
For example, ~\citet{Biderman2023PythiaAS} controlled for training variables including model architecture, training scale, model checkpoints, hyper-parameters, and source libraries to investigate the influence of data frequency on ICL and bias behaviour in LLMs. Another key challenge is the absence of benchmark datasets that effectively justify the causal effects of the investigating factors on ICL~\cite{Zhang2023UnderstandingCW,Jin2023CanLL}.
One possible solution is to generate synthetic datasets with domain-specific expertise and develop methods in causal discovery and inference ~\cite{Swaminathan2023SchemalearningAR, Kcman2023CausalRA} for interpreting ICL through a causal lens.
\vspace{-1mm}
\paragraph{Evaluation}
Current research efforts typically measure ICL by assessing task performance or optimizing criteria such as gradient~\cite{Oswald2022TransformersLI} and token loss~\cite{Olsson2022IncontextLA} during the pre-training stage.
However, \citet{Schaeffer2023AreEA} have recently argued that emergent abilities (e.g., ICL) discussed in some prior studies appear to be mirages, due to the researchers' choice of evaluation metrics.
Their hypothesis is that choosing a metric that nonlinearly or discontinuously deforms per-token error rates and the limited size of test datasets may not provide an accurate estimation of the performance of smaller models. The core challenge is that aggregated performance metrics do not adequately assess ICL ability across various scenarios or predict their behaviour in new tasks alongside data distributions drifting.
Dedicated criteria explicitly designed for the assessment of ICL are currently lacking.
In addition, the evaluation process typically encompasses multiple models with a distinct objective in existing LLM paradigms, such as reinforcement learning from human feedback (RLHF)~\cite{Christiano2017DeepRL}.
A potential solution involves identifying and addressing specific capability gaps to enhance predictions of model performance on novel tasks.
For example, \citet{Burden2023InferringCF} incrementally inferred the capability of LLMs with subsets of tasks before assessing more complex dependencies. 
\vspace{-1mm}
\paragraph{Demonstration Selection}
While research exploring the relationship between demonstration and ICL is expanding~\cite{Min2022RethinkingTR,xiang2024addressing}, it is mostly limited to interpretation based on a finite number of demonstrations. 
The core challenge lies in the exponential growth of possible demonstrations with the increase in examples.
A potential approach is to formalize the demonstration selection as a sequential decision-making problem~\cite{Zhang2022ActiveES}, aiming to learn an approximation of the expected reward from demonstrations to identify those most beneficial for control experiments.
Other potential solutions can include disentangling features supportive for performing ICL on downstream tasks, and controlling for these features in demonstrations~\cite{Si2023MeasuringIB}.
\paragraph{Trustworthiness}
Trustworthiness issues such as fairness, truthfulness, robustness, bias, and toxicity are significant concerns for LLMs.
Exploring these properties within ICL in LLMs is particularly challenging due to their unanticipated nature~\cite{Kenthapadi2023GenerativeAM}.
It is difficult to analyze the relationship between various aspects of ICL and factors relating to trustworthiness when LLM training objectives and downstream tasks are inconsistent. 
Safety concerns have also become one of the most pressing issues. Studies~\cite{Perez2022IgnorePP,Bai2022TrainingAH} have shown that LLMs can be manipulated to perform harmful and dangerous ICL through exposure to toxic demonstrations. Understanding ICL could play a crucial role in addressing the trustworthiness and safety issues associated with LLMs. For example, knowledge of how LLMs incorporate biases during ICL can guide the development of debiasing models. Furthermore, insights gained from studying how LLMs respond to toxic demonstrations can inform the design of countermeasures aimed at detecting and filtering out harmful input.




\vspace{-1mm}
\section{Conclusion}
\vspace{-1mm}
This paper presents a comprehensive review of current research efforts focused on interpreting and analyzing ICL in LLMs.
We organize these advancements into theoretical and empirical perspectives, highlighting existing challenges and discussing potential avenues for further research in this area.
We believe this survey will serve as a valuable resource for encouraging further exploration into the interpretation of ICL of LLMs.


\section*{Limitation}
While we referenced numerous studies to interpret and analyze in-context learning, many of them are only briefly described due to space limitations.
Our aim was to provide an overview of existing research efforts into ICL interpretation, and to organize previous research within a principled framework.
Moreover, the survey primarily focuses on the ICL ability, which has been extensively investigated in previous studies.
Nevertheless, there are other intriguing capabilities that have emerged in LLMs, such as chain-of-thought~\cite{Chu2023ASO,Feng2023TowardsRT} and instruction following~\cite{ Wei2021FinetunedLM,Chung2022ScalingIL,Ouyang2022TrainingLM}, which are not included in this survey.

\section*{Acknowledgements}
This work was supported in part by the UK Engineering and Physical Sciences
Research Council (EPSRC) through a Turing AI Fellowship (grant no. EP/V020579/1,
EP/V020579/2) and a New Horizons grant (grant no. EP/X019063/1), and Innovate UK through the Accelerating Trustworthy AI programme (grant no. 10093055).

\bibliography{anthology,custom}


\clearpage

\end{document}